\documentclass{article}

\usepackage{microtype}
\usepackage{graphicx}
\usepackage{subfigure}
\usepackage{booktabs} 

\usepackage{hyperref}

\usepackage[accepted]{tom2023}

\usepackage{amsmath}
\usepackage{amssymb}
\usepackage{mathtools}
\usepackage{amsthm}

\usepackage[capitalize,noabbrev]{cleveref}

\theoremstyle{plain}

\theoremstyle{definition}

\theoremstyle{remark}

\usepackage[textsize=tiny]{todonotes}

\icmltitlerunning{Meta Neural Coordination}

\begin{document}

\twocolumn[
\icmltitle{Meta Neural Coordination}




\begin{icmlauthorlist}
\icmlauthor{Yuwei Sun}{*}
\end{icmlauthorlist}

\icmlaffiliation{*}{The University of Tokyo}

\icmlcorrespondingauthor{Yuwei Sun}{ywsun@g.ecc.u-tokyo.ac.jp}


\vskip 0.2in
]

\printAffiliationsAndNotice{}




\begin{abstract}
Meta-learning aims to develop algorithms that can learn from other learning algorithms to adapt to new and changing environments. This requires a model of how other learning algorithms operate and perform in different contexts, which is similar to representing and reasoning about mental states in the theory of mind. Furthermore, the problem of uncertainty in the predictions of conventional deep neural networks highlights the partial predictability of the world, requiring the representation of multiple predictions simultaneously. This is facilitated by coordination among neural modules, where different modules' beliefs and desires are attributed to others. The neural coordination among modular and decentralized neural networks is a fundamental prerequisite for building autonomous intelligence machines that can interact flexibly and adaptively. In this work, several pieces of evidence demonstrate a new avenue for tackling the problems above, termed Meta Neural Coordination. We discuss the potential advancements required to build biologically-inspired machine intelligence, drawing from both machine learning and cognitive science communities.
\end{abstract}

\section{Introduction}

Common sense is more than just a collection of facts, but a set of mental models that allow individuals to understand and navigate the world around them \cite{minsky}. Cognitive science research has shown that specialized modules in the human brain coordinate to support complex cognitive processes. The Thousands Brains theory \cite{brains} suggests that the neocortex learns complete models of objects through many models of each object distributed throughout cortical columns of the neocortex. However, conventional deep neural networks lack sufficient modularity and sparsity. For example, ChatGPT \cite{chat} is built upon large-scale training and a monolithic Transformer architecture. A limited degree of alignment between its beliefs and those of a human user is performed via Reinforcement Learning from Human Feedback (RLHF) \cite{rlhf}. As a result, we have been observing many edge cases, false facts, and misuses of the model.

The Future of Life Institute has called for a six-month moratorium on the development of AI research to allow AI companies and regulators time to formulate safeguards to protect society from potential risks of the technology. Furthermore, there have been calls and joint workshops to reunite machine learning research with other scientific disciplines, including neuroscience, cognitive science, mathematics, and psychology \cite{neuroai,juliani2022on}. In this thought piece, we propose a potential avenue to address problems in machine learning with inspiration from the Theory of Mind and Consciousness. We argue that both a diverse collection of neural modules and an efficient interface for communication and coordination among modules are missing from current neural network architecture.

Neural coordination involves organizing neural modules of different functionalities. By understanding how neural coordination works among artificial neural modules, we can develop better models of how the brain operates and design more effective learning systems. One analogy to the coordination among neural modules is the Global Workspace Theory (GWT) \cite{Baars1988,VANRULLEN2021692}. In GWT, multiple neural networks cooperate and compete in solving problems via a communication bottleneck for information sharing. Using different kinds of metadata about individual neural networks, such as measured performance and learned representations, shows potential to learn, select, or combine different learning algorithms to efficiently solve a new task. The integration of knowledge representations from these different neural modules enables the process of reasoning and planning. GWT is one of the most promising theories to understand consciousness and build the interface for communication among neural modules.

This work reviews and discusses several studies regarding neural coordination for improving conventional deep neural networks. These studies demonstrate that neural coordination improves model performance in tackling out-of-distribution (OOD) and non-independent and identically distributed (non-IID) data. We urge the community to investigate the neural coordination problem by integrating machine learning, neuroscience, and cognitive science to advance the development of intelligent machines that learn and adapt over a lifetime through inter-module communication.

\section{Meta Neural Coordination}

The study of neural coordination in modular and decentralized neural networks focuses on how a meta-observer can accomplish tasks by utilizing the models constructed by other individual neural modules. In this section, we present different approaches to developing the theory of mind in conventional neural networks and demonstrate that optimizing the communication and collaboration among these neural modules can significantly enhance the learning system's performance and adaptability in unseen tasks.

\subsection{Learning from Replica Neural Modules with Diverse States}

To build a general learning system that efficiently allocates resources for different tasks, it is essential to possess the ability to consider others' perspectives for both cooperation and competition. To achieve this, a collection of individual neural modules acting as replicas of a prototype model aim to learn the overall data distribution of the environment by constructing different world models and sharing the learned knowledge among modules through communication. In this regard, decentralized neural networks are designed to facilitate knowledge transfer between neural modules trained on separate local data. The goal is to learn a global model that can generalize to unseen situations through local model sharing and knowledge aggregation \cite{survey}.

One of the main practical challenges in coordinating decentralized neural networks is tackling out-of-distribution (OOD) and non-independent and identically distributed (non-iid) data. Non-iid data refers to situations where data samples across local models are not from the same distribution, making it difficult to transfer knowledge between them. OOD data, on the other hand, refers to inputs that have domain discrepancies with specific styles, often due to differences in the data collection environment. For example, an autonomous vehicle that learns to drive in a new city might leverage the driving data of other cities learned by different vehicles. Since different cities have different street views and weather conditions, it would be difficult to directly communicate and share the knowledge learned by these models. These challenges make it necessary to develop approaches to enabling effective coordination and knowledge transfer among decentralized neural modules.

This problem is closely related to the fields of transfer learning and domain adaptation, which study distribution shifts and negative transfer that hinder a model's generalization to unseen tasks. To address this issue, recent work such as Federated Knowledge Alignment (FedKA) \cite{sun2022fedka} proposed using a shared global workspace to align knowledge representation among neural modules. FedKA consists of three components, a feature disentangler, embedding matching, and federated voting, which aim to improve the transferability of knowledge representations and reduce the inefficiency in neural module communication.

\subsection{Building the Hierarchy of Neural Modules for Different Functionalities}

Hierarchical neural networks consist of multiple neural modules connected in a form of an acyclic graph. One of the key advantages of hierarchical neural networks is their ability to decompose complex tasks into simpler sub-tasks, which can be efficiently handled by individual neural modules. The conscious prior theory \cite{bengio2019} proposed such a theoretical framework of a sparse factor graph to learn module relations in the mapping of high-level semantic variables. Homogeneous Learning \cite{homo} is a hierarchical neural network approach that aims to tackle a task in sequential actions, by selecting the optimized module at each time step and recursively updating a learning policy. A meta in Homogeneous Learning observes the states of itself and its surrounding environment (other modules), computing the expected rewards for taking different actions of communicating states. With a model of external reality and possible actions, the meta can try out various alternatives and conclude which is the best action \cite{nature}. Then, the optimized learning policy allows a more efficient adaptation of the hierarchical system to new tasks by enabling better planning and leveraging of different neural modules.

Moreover, the use of a meta observer and the hierarchical organization of neural modules is closely related to the concept of System 1 and System 2 AI \cite{system12}. System 1 processing is fast and intuitive, relying on local specialized networks. On the other hand, System 2 processing that selects from the bottom-up System 1 inputs, is slow and explicit, relying on effortful cognitive processes that require more distributed processing of neural modules and flexible interactions between them.

\subsection{Leveraging Multi-modal Modules for Improved Communication Richness}

Recent work \cite{mental} suggests a measurement approach to the ineffability incurred during the mental representation and ascription of thoughts, beliefs, and desires to others. Leveraging multi-modal sensor information \cite{clip,dalle,dalle2,chat} can improve the richness of module communication and obtain refined cross-modal representations that can be potentially reused for different downstream tasks. In this regard, information in the real world often comes in different modalities, and degeneracy \cite{baby} refers to the ability of multiple configurations of neural modules to carry out a single function. The degeneracy in multi-modal module communication creates redundancy and improved richness of information, allowing the system to function even with the loss of one modality. 

Self-supervised learning has emerged as a promising approach to coordinating among multi-modal neural modules and obtaining rich communication states. Unlike supervised learning, self-supervised learning learns by observing relevant and irrelevant modality information instead of using hard labels for training. For instance, CLIP \cite{clip} computes a cosine similarity matrix among all possible candidates of images and texts within a batch to obtain cross-modal representations. Similarly, SimCLR \cite{simclr} and BYOL \cite{byol} are other popular self-supervised learning methods that leverage the contrastive learning objective to learn useful feature representations across different modalities. Moreover, Question-Image Correlation Estimation (QICE) \cite{zhu20} is a self-supervised method that trains on relevant image and question pairs to tackle Visual Question Answering tasks. Learning by observation facilitates modeling human-like cognitive processes, the ability to reason and communicate based on multi-modal sensory inputs.

\section{Discussion}
There are several exciting avenues to explore in the field of neural coordination for understanding how different modules can work together to learn and adapt over time. We offer several preliminary ideas for future work here.

\subsection{Neural Coordination in Transformer Models}

The study of neural coordination in Transformer models involves the implementation of mixture of experts \cite{scale,vit,sparse} that allows multiple independent neural modules to form a shared workspace. Then, a routing function allows a sparse communication between a few of these modules. The selective communication and coordination would be critical for overcoming the problem of catastrophic forgetting, i.e., the tendency of neural networks to forget previously learned knowledge when presented with new information. By dividing the network into modular components, it is possible to allow different parts of the network to learn and adapt without disrupting the knowledge that other parts of the network have learned. We have seen several recent efforts in this avenue \cite{sparse,outrage,coordinate}.

\subsection{Learning Discrete Communication with Associative Memory Attractors}

To better coordinate among neural modules, it is assumed that implementing representation discretization of the learned knowledge of the expert modules can be helpful, as various brain areas are tuned to discrete variables while deep neural networks rely on continuous representations. There are several approaches for building discrete representations, such as vector quantization (VQ) \cite{vq,coordinate} and associative memory \cite{hopfield}. Notably, in methods of associative memory, an observed state converges to a fixed attractor point close to one of the stored learnable patterns from previous tasks in the long-term memory. The Hopfield network is one type of associative memory, with a more recent modern Hopfield network proposed by Krotov and Hopfield \cite{modernhf} and subsequently further developed by Demircigil et al. \cite{demi}. Moreover, the most recent continuous Hopfield network \cite{hficlr} demonstrated the mathematical formulation of the energy function that underpins the attention mechanism in Transformer models. It has shown the ability to control the attraction basins of the individual patterns and the formation of metastable states with an inverse temperature $\beta$. In addition, the continuous Hopfield network could greatly increase the memory capacity for storing the discrete communication states of neural module coordination.

\section{Conclusion}
Meta Neural Coordination offers a new approach to tackling the challenges of uncertainty and adaptability in deep learning, drawing inspiration from mental state representation in the theory of mind. Coordinating neural modules by representing their internal states and facilitating efficient knowledge sharing among different modules in the shared global workspace, enables swift adaptation to unseen tasks over a lifetime. This work highlights the potential for building autonomous and flexible machine intelligence by obtaining understanding from machine learning and cognitive science. 

\nocite{langley00}

\bibliography{example_paper}
\bibliographystyle{tom2023}

\end{document}